\documentclass{article}
\usepackage{spconf,amsmath,graphicx,bm}
\usepackage{setspace}

\title{
Data Augmentation for Low Resource Sentiment Analysis using Generative Adversarial Networks
}
\name{Rahul Gupta}
\address{Amazon.com, USA}
\begin{document}
\ninept
\maketitle
\begin{abstract}

Sentiment analysis is a task that may suffer from a lack of data in certain cases, as the datasets are often generated and annotated by humans.
In cases where data is inadequate for training discriminative models, generate models may aid training via data augmentation. 
Generative Adversarial Networks (GANs) are one such model that has advanced the state of the art in several tasks, including as image and text generation. 
In this paper, I train GAN models on low resource datasets, then use them for the purpose of data augmentation towards improving sentiment classifier generalization. 
Given the constraints of limited data, I explore various techniques to train the GAN models.
I also present an analysis of the quality of generated GAN data as more training data for the GAN is made available. 
In this analysis, the generated data is evaluated as a test set (against a model trained on real data points) as well as a training set to train classification models. 
Finally, I also conduct a visual analysis by projecting the generated and the real data into a two-dimensional space using the t-Distributed Stochastic Neighbor Embedding (t-SNE) method.
 
\end{abstract}


\vspace{-2mm}
\section{Introduction}
\vspace{-2mm}
\label{sec:intro}
Since their introduction, Generative Adversarial Networks (GANs) \cite{goodfellow2014generative} have established themselves as powerful models, outperforming other generative models in the tasks such as image  and text generation \cite{yu2017seqgan, denton2015deep}.
Variations of GANs, such as Wasserstein GAN \cite{arjovsky2017wasserstein}, coupled GAN \cite{liu2016coupled} and StackGAN \cite{zhang2017stackgan}, have been proposed to improve upon the GAN architecture for various tasks.
Recently, the application of GANs has also been extended to affective computing \cite{han2018adversarial} with applications to emotion transformation in images \cite{ding2017exprgan}, image and video generation with emotional attributes \cite{bao2018towards} and emotional data augmentation \cite{pham2018generative}.
In this work, I apply GAN models to another critical task: low resource sentiment analysis. 
Although several sentiment analysis tasks come with sufficient amount of data to train complex low-error models, large quantities of data may not be available on newly formed sentiment analysis tasks or other tasks with constrained resources. 
In order to improve the classification performance on such tasks, I experiment with GAN models as a channel to synthetically generate additional data-points. 
To achieve this, I propose a variation of the conditional GAN (cGAN) model \cite{goodfellow2014generative} to generate data on low resource datasets. 
I conduct further analysis to understand value added by appending the cGAN generated data. 

{\bf Previous Work:} Sentiment analysis \cite{pang2008opinion} is a classical problem to evaluate the affective states associated with a human opinion or a reaction given to a given event.
Application of machine learning techniques for sentiment analysis is a widely researched field. 
Survey articles such as ones by Liu et al. \cite{liu2012survey} and Medhat et al. \cite{medhat2014sentiment} provide a summary of such techniques. 
Sentiment analysis has been previously studied in low resource settings by application of methods such as transfer learning \cite{gupta2018semi} and semi-supervised learning \cite{goldberg2006seeing}.
The set of techniques proposed for sentiment analysis in absence of labeled data include manifold regularization \cite{gupta2018semi}, semi-supervised recursive autoencoders \cite{socher2011semi}, document word co-regularization \cite{sindhwani2008document} and latent variable models \cite{tackstrom2011semi}. 
On the other hand, GAN models were proposed in 2014 and a tutorial by Goodfellow \cite{goodfellow2016nips} provides a background on GANs.
Jing et al. \cite{han2018adversarial} provide a summary on the application of adversarial Training in Affective Computing and Sentiment Analysis.
I pursue a data augmentation approach to aid sentiment analysis in this paper.
Data augmentation through GANs has been used in other tasks such as enhancing emotion recognition through speech \cite{sahu2018enhancing, sahu2018adversarial}, human pose estimation \cite{peng2018jointly} and medical image synthesis \cite{shin2018medical}. 
In my work, I use a variant of cGAN and apply various heuristics to achieve their convergence.
This is the first work that evaluates and analyzes GAN models as data augmentation tools for sentiment analysis. 

In order to train cGAN models that can augment data samples for a given low resource task at hand, I propose a variant of cGANs that also uses a baseline classifier trained on the task of interest.
I apply several other tricks to obtain convergence on the cGAN model training (such as model pre-training, noise addition to the inputs and one sided label smoothing).
I evaluate the thus trained cGAN models on two different low resource sentiment analysis tasks: (i) movie review and, (ii) product review. 
Although, I observe that the data generated using such cGAN models can by itself be used to train discriminative models, they provide a marginal gain when appended to the rest of the available real data. 
On the movie and product review datasets, I obtain relative improvements of 1.6\% and 1.7\% (respectively) against a baseline model only using the real data.
I conduct further analysis on the improvements yielded by data augmentation as more data is made available for training the cGAN model. 
To this end, I use a larger publicly available sentiment analysis data from a social media platform.
I train cGAN models by incrementing data available during training and observe the gains in sentiment classification. 
Finally, I also perform a data distribution analysis (using t-SNE embedding), comparing the distribution of cGAN generated data against real dataset.  
In the next section, I describe the cGAN architecture used in my experiments, followed by a description of the low resource datasets. 

\vspace{-2mm}
\section{Conditional GAN setup for data augmentation}
\vspace{-2mm}
\label{sec:method}
In this section, I describe the setup for using cGANs in augmenting data for sentiment classification.
First, I describe the architecture of the cGAN models trained on low resource sentiment analysis datasets.
Followed by this, I describe the training methodology and tricks for the cGAN architecture. 

\vspace{-1mm}
\subsection{cGAN architecture}
\vspace{-1mm}

\begin{figure}
\centering
\includegraphics[scale=0.35]{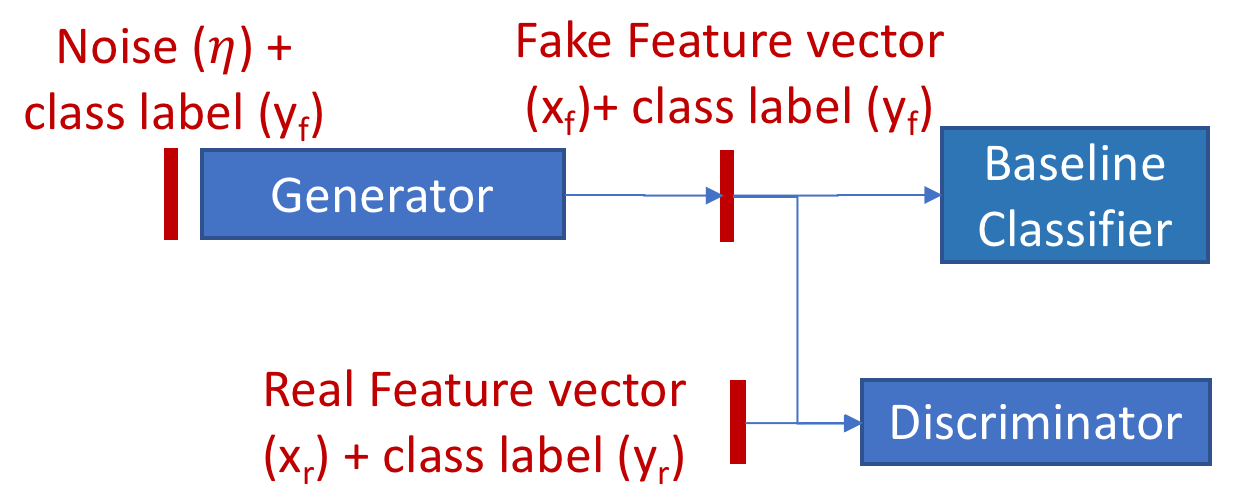}
\vspace{-3mm}
\caption{cGAN architecture and notations used in my experiments. The generator parameters are updated based on signals from discriminator and the baseline classifier.}
\vspace{-3mm}
\label{fig:gan}
\end{figure}

I use a cGAN architecture as shown in Figure~\ref{fig:gan}, inspired by similar cGAN variants \cite{denton2016semi} except for the addition of a previously available baseline classifier (along with the generator and discriminator components). Below, I briefly describe the three cGAN components. \\

\noindent{\bf Generator}: The generator is a feed-forward neural network with inputs as a noise vector $\bf \eta$ along with a one hot encoding of sentiment class $\bf y_f$ (e.g., in a binary sentiment classification task $\bf y_f$ would be a two dimensional vector). 
Given these inputs, the generator is expected to produce a feature vector $\bf x_f$ (a Doc2Vec representation \cite{dai2015document} of sentences in my experiments), which corresponds to the class encoding $\bf y_f$ input to the generator. \\

\noindent{\bf Discriminator}: Given pairs of real feature vectors $\bf x_r$ and the associated class vector $\bf y_r$ (again in the form of a one hot encoding) along with the fake feature vectors $\bf x_f$ produced by the generator as per the class encodings $\bf y_f$, the discriminator attempts to classify fake pairs $[\bm x_f, \bm y_f]$ against the real pairs of data-points $[\bm x_r, \bm y_r]$. 
In my experiments, the discriminator is also a feed-forward neural network which accepts inputs as feature vectors concatenated with associated class embeddings and outputs the probabilities of samples being real. \\

\noindent{\bf Baseline classifier}: Apart from generator and discriminator as in a standard cGAN, I also append a baseline classifier trained previously on the dataset available for the task at hand. 
In low resource settings, this classifier is to be trained on the limited amount of data available.
Hence, I chose a shallow neural network as the architecture for the baseline classifier, as it contains fewer parameters to be optimized.  
The objective of appending a baseline classifier in my cGAN architecture is, given an input class encoding $\bf y_f$, the generator is encouraged to produce a corresponding sample $\bf x_f$ with high confidence as is adjudged by the baseline classifier.
I describe the cGAN training loss functions and the tricks I use to train the cGAN model in more details below. \\ 

\vspace{-1mm}
\subsection{Training cGAN model}\label{sec:gan}
\vspace{-1mm}
The cGAN model is trained using an algorithm where discriminator and generator parameters are tuned alternately and iteratively.
I do not update the baseline classifier parameter during cGAN training.
During an iteration, discriminator parameters are tuned to minimize the cross entropy loss ($L_D$) as defined in equation~\ref{eq:d_loss}.
$[\bm x_f; \bm y_f]$ represents a vector concatenation of $\bm x_f$ and the associated label representation $\bm y_f$.
$D([\bm x_r; \bm y_r])$ and $D([\bm x_f; \bm y_f])$ are the probabilities assigned by the discriminator to the pairs $[\bm x_r, \bm y_r]$ and $[\bm x_f, \bm y_f]$ being real, respectively.
$y$ is 1 for real data-points $\bm x_r$ and 0 for the fake data-points $\bm x_f$.

\begin{equation}\label{eq:d_loss}
L_D = - y \log(D([\bm x_r; \bm y_r])) -  (1-y) \log(1-D([\bm x_f; \bm y_r])) 
\end{equation} 

After updating the discriminator parameters, I tune the generator parameters using the loss function $L_G$ in equation~\ref{eq:g_loss}. 
Apart from the standard generator loss $L_{G1}$ to fool the discriminator, I also add another cross entropy loss $L_{G2}$ between the class prediction returned by the baseline classifier for $\bf x_f$ with respect to the expected class label $\bf y_f$.
The objective of this part of the loss function is to encourage the generator to produce $\bf x_f$ such that the baseline classifier is in agreement with the association between $\bf x_f$ and $\bf y_f$.

\begin{equation}\label{eq:g_loss}
L_G = L_{G1} + \lambda L_{G2} 
\end{equation} 

\begin{equation}\label{eq:g_loss_pw}
\begin{aligned}
{\text{Where,}\;\;}L_{G1} = - \log(D([\bm x_f; \bm y_f])); \bm x_f = G(\bm \eta) \\
L_{G2} = - \text{CE}(\bm y_f, C(\bm x_f)) 
\end{aligned}
\end{equation} 

In equation~\ref{eq:g_loss_pw}, $G(\bm \eta)$ represents the output ($\bm x_f$) yielded by the generator when the input is a noise vector $\bm \eta$. 
$C(\bm x_f)$ is the prediction probabilities returned by the baseline classifier on the sample $\bm x_f$ for the sentiment classes and $\text{CE}(\bm y_f, C(\bm x_f))$ is the associated cross-entropy.
$\lambda$ is a hyper-parameter to tune relative weights between $L_{G1}$ and $L_{G2}$.
I note that the generator from a trained cGAN does implicitly learn the relationship between $\bf x_f$ and $\bf y_f$.
The generator is targeted to produce $\bf x_f$ and $\bf y_f$ pairs such that the discriminator can not distinguish the fake pairs against a real pair of $\bf x_r$ and $\bf y_r$.
I hypothesize that the explicit addition of $L_{G2}$ helps the generator to generate fake pairs for which it is more confident of the association between $\bf y_f$ and $\bf x_f$, particularly in a low resource setting.  
Apart from the addition of the baseline classifier, I also use tricks such as addition of data points from an external dataset, noise addition to the inputs as well as one sided label smoothing to achieve better convergence of the cGAN losses.
I briefly describe these tricks below. \\ 

\noindent{\bf Initialization with other dataset:} Since my target tasks come with a limited amount of training data, I pre-train the cGAN model on a larger external dataset from a related task.
The large dataset helps the cGAN model to converge to relatively stable parameters, which can then be fine tuned on the smaller dataset available for the task at hand. 
Since I use an external dataset for pre-training the cGAN models, I hypothesize that increasing $\lambda$ towards the final few iterations of cGAN optimization can help the generator to tune better to the low resource dataset. \\

\noindent{\bf Noise addition:} After pre-training the cGAN model, I add a Gaussian random noise to the in-domain real input feature vectors $\bf x_f$ for each iteration of cGAN training.
The injected Gaussian noise carries a zero mean and a diagonal covariance matrix with values 0.02. 
Adding noise to inputs is another regularization method for GANs \cite{goodfellow2016nips} and prevents the cGAN from over-fitting to the smaller dataset. \\ 

\noindent{\bf One sided label smoothing:} Another trick that I found particularly useful in my experiments was one sided label smoothing \cite{salimans2016improved}.
It also acts as a regularizer and prevents providing large gradients to the generator.
Hence, the generator parameters do not differ by a large value after pre-training.\\

Apart from the above tricks, I use batch normalization and randomly training the generator multiple times in each cGAN training iteration to achieve better quality fake samples.
The training batch is normalized to carry zero mean and unit variance per feature dimension (this implies that the noise added to $\bm x_r$ carries a strength of 2\%, against the signal strength). 
In the next section, I describe the datasets I use in my experiments. 

\vspace{-2mm}
\section{Datasets}
\vspace{-2mm}
My experiments are primarily geared towards two sentiment classification tasks: (i) Movie review and, (ii) Product review. 
The datasets for the two tasks are described in more detail below:\\ 

\noindent{\bf Movie review dataset:} The first dataset used to train cGAN models for my experiments is the movie review dataset \cite{pang2004sentimental}. 
Each review in the dataset comprises of multiple sentences, along with an associated positive/negative sentiment annotation.
The dataset consists of $\sim$2k samples and I perform a 50:50 random split to define training and testing portions on the dataset.\\ 

\noindent{\bf Product review dataset:} The product review dataset \cite{kotzias2015group} consists of reviews on Amazon, yelp or IMDB.
The dataset consists of $\sim$3k samples, annotated with positive/negative sentiment label.
I split the dataset into training and testing portions of equal size.\\ 

The cGAN model (as well as the baseline classifier used in the cGAN model) for each dataset is trained using the associated training split. 
I conduct a classification evaluation on the testing portion, as described in the next section.
I use the Twitter dataset \cite{go2009twitter} consisting of $\sim$1.6M tweets to pre-train the cGAN model. 
Each tweet is annotated with a positive/negative sentiment, also yielding a two dimensional $\bm y_r$ for pre-training. 
The feature representation $\bf x_f$ used in my experiments is a Doc2Vec \cite{dai2015document} representation, trained on the Wikipedia corpus.
I note that the cGAN models trained in my experiments are geared towards mimicking the Doc2Vec representations of the sentences, as opposed to more recent efforts around training GANs to obtain sentences themselves \cite{yu2017seqgan}.
This is an avenue I will consider for future research. 
The feature and label representations $\bf x_f, \bf y_f$ are common in between the two primary datasets for investigation and the Twitter dataset. 
This allows for a pre-training followed by fine-tuning without altering the cGAN architecture.

\vspace{-2mm}
\section{Experimental setup}
\vspace{-2mm}
\label{sec:exp_setup}
I initially train separate cGAN models for movie and review datasets, using the scheme described in section~\ref{sec:gan}.
Once trained, I generate 10k fake pairs of feature samples ($\bf x_f$) and class vectors ($\bf y_f$) from the cGAN generator.
This fake data is then used to train another classifier $C_f$. 

Apart from the baseline classifier $C_b$ and classifier trained on the cGAN data $C_f$, I also train a classifier $C_t$ on the twitter dataset.
Through $C_t$, I aim to estimate the performance obtained using a transfer learning approach. 
In the next section, I describe my evaluation methodology.

\begin{table}
\centering
\caption{Classification accuracies obtained on the test partitions of movie review and UCI dataset using the various classifiers. Performance yielded by $C_f$ are significantly better than chance (using a binomial proportions test at $<$5\% significance level.)} 
\begin{tabular}{ l|cc }
\hline
Classifiers used & UCI dataset & Movie review\\
Chance & 50.0 & 50.0 \\
$C_b$ & 63.3 & 72.0 \\
$C_f$ & 56.3 & 55.7 \\
$C_t$ & 63.9  & 62.6\\
$C_b, C_f$ & 64.3 & 73.2 \\ 
$C_b, C_f, C_t$ & 64.5 & 74.0 \\ \hline 
\end{tabular}
\label{tab:results2}
\vspace{-5mm}
\end{table}

\vspace{-1mm}
\subsection{Results}
\vspace{-1mm}
For each of the three classifiers $C_b, C_f, C_t$, I report the accuracy achieved on the test partition of the target datasets.
I also report accuracies on a bagged approach, where I combine confidence scores from either a subset of or all three classifiers before the class assignment.
The combination weight is computed based on a small held-out set from the training portion (we just need to tune three weights, so a small validation carve out is sufficient).
Table~\ref{tab:results2} reports the accuracies.

From the results, I observe that the accuracies yielded by the classifier $C_f$ trained on the cGAN generated data is significantly higher than chance.
This implies that the model trained on the fake data carries discriminative power.
Although the $C_f$ accuracies are not as high as the classifier $C_b$, when bagged together, I obtain better accuracies on both the datasets. 
I also observe that the accuracies yielded by the classifier $C_t$ to be higher than chance, indicating that a transfer of knowledge from the twitter dataset is possible in the other two datasets of interest. 
Finally, the best accuracies are obtained by bagging across all the three classifiers.

\vspace{-2mm}
\section{Analysis on the data generated using cGAN model}
\vspace{-2mm}
Results in the previous section show that the data generated using cGAN models provide marginal improvements in classification performance in low resource settings.
I perform further analysis to assess the impact of data size in training cGAN models, as well as a visual assessment of the generated fake data. 
Using the twitter dataset, I perform two sets of analysis: (i) evaluating the impact of the training dataset size on the cGAN model and, (ii) a data distribution analysis using t-SNE projections. 
I discuss them in detail below.

\vspace{-1mm}
\subsection{Evaluating the impact of the dataset size on cGANs}
\vspace{-1mm}
\label{sec:evaluation}
I evaluate the impact of the size of the available dataset on the cGAN model in this section.
The quality of the cGAN data generated is evaluated in two settings: when used as a training set as well as a test set. 
The setup of my analysis is as follows.
Initially, I segment the twitter dataset into a training (~600k samples) and test set (~1M samples).
I train a neural network classifier $C_b^{\text{full}}$ on the entire training dataset.
In addition, a cGAN model is trained on a subset of $N$ samples from the training set ($N = 500, 1k, 2k, 4k, 8k$).
I sample 10k fake feature and class vectors pairs ($\bm x_f, \bm y_f$) from the trained cGAN and use them as training and testing sets as described below. \\ 

\begin{figure}
\centering
\includegraphics[scale=0.4]{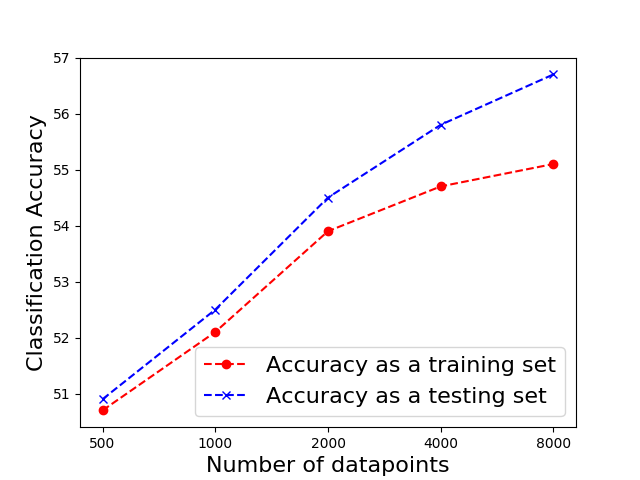}
\caption{Plot showing accuracies on the generated data when used as a test set as well as accuracy yielded by a model trained using the generated data on a real test set.} 
\vspace{-3mm}
\label{fig:gan_acc}
\end{figure}

\noindent{\bf As a test set:} 
In this setting, I use the fake data pairs $\bm x_f$ and $\bm y_f$ generated using cGAN as a test set.
This setting is same as the one during cGAN training where the generated data is fed to the baseline classifier to obtain the cross-entropy loss $L_{G2}$. 
For a given $\bm x_f$, I treat $\bm y_f$ as the ground truth.
Given a set of $N$ training samples, I train the cGAN model described in section~\ref{sec:gan}.
The baseline classifier used in the cGAN models is also trained on the available set of $N$ training samples.
The classifier $C_b^{\text{full}}$ trained on the entire training set is then evaluated on the generated fake data. 
A better accuracy on the fake data as test set implies a better trained cGAN as the loss $L_{G2}$ in equation~\ref{eq:g_loss} is expected to be lower.
However, it may not directly imply a higher quality of the generated data when used to augment sparse datasets. 
The accuracies on the fake samples as I increase $N$ for cGAN training is shown in Figure~\ref{fig:gan_acc}.\\

\noindent{\bf As a training set:}  
In this setting, I use data generated using cGAN model as a training dataset (akin to the experiment in Section~\ref{sec:exp_setup}).
I train a shallow NN on the fake data with $\bm x_f$ as input features and $\bm y_f$ as target labels. 
The NN is then evaluated on the real test set of $\sim$1M samples.
The accuracy yielded by this NN on the real test set reflects the direct value of the generated data in enhancing classification through data augmentation. 
Figure~\ref{fig:gan_acc} shows the accuracy achieved on the test set as I increase $N$ for the cGAN training. 

Figure~\ref{fig:gan_acc} suggests that the accuracy of the fake data as test set is higher than when used as a training set for all values of $N$.
Moreover, as $N$ increases, accuracy on the fake data as test set increases more rapidly than when it is used as a training set.
This implies that as $N$ increases, the loss $L_{G2}$ is expected to decrease and the overall loss value for cGAN is expected to be lower. 
However, performance due to data augmentation does not increase commensurately. 
Also, the figure suggests that the gain yielded by data augmentation saturates as $N$ increases.
This hints towards an upper limit to the gains yielded by the cGAN model. 
Overall, I observe that there is a discrepancy in the expected loss decrease in cGAN model training and the expected increase in performance due to data augmentation.
I conduct further analysis on the cGAN generated data using t-SNE and observe plausible reasons for this discrepancy. 
 
\subsection{Analysis of the generated dataset using t-SNE}

\begin{figure}
\centering
\includegraphics[scale=0.45]{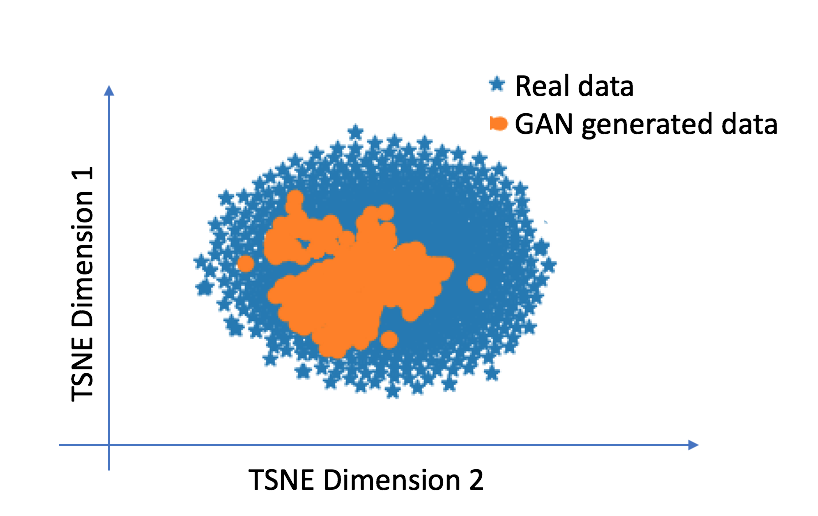}
\vspace{-3mm}
\caption{Real and fake data distribution, as observed on a 2-D projection of data-points obtained using the t-SNE method.}
\label{fig:gan_tsne}
\end{figure}

In order to further understand the distribution of the data generated using cGAN models, I project the data into a lower dimension for visual analysis. 
Given data-points from the real dataset and fake data generated using the cGAN, I use the t-SNE method to project the data into a two dimensional subspace.
I sub-sample 2k data-points from real dataset as well as fake data generated by the cGAN model trained using 8k samples in the previous section. 
The t-SNE projection parameters are obtained on the combination of the fake and real data-points and Figure~\ref{fig:gan_tsne} shows the distribution of fake and real data-points.
I observe that while the fake data overlaps with the real data points, it does not completely cover the subspace spanned by real dataset.
This indicates that cGAN is generating data-points in a limited region of the actual distribution of real data-points.
Despite the application of tricks mentioned in Section~\ref{sec:gan}, I was not able to obtain a fake data distribution that completely resembles the distribution of real data.
This problem is akin to mode collapse in GAN models, where a GAN model tends to produce data samples in a limited region of the feature space. 

It is also plausible that the discrepancy in using the fake data as training and testing sets in Section~\ref{sec:evaluation} could be explained by this difference in the distribution of real and fake data.
While fake data may lie in a region of feature space that could be well classified by the baseline classifier, a classifier trained on the fake data can not correctly classify test instances drawn from a region not covered by the fake samples. 
I expect that addressing the discrepancy in the coverage of data distribution can lead to further improvement of classification. 
I aim to address this as a part of future research.

\vspace{-2mm}
\section{Conclusion}
\vspace{-2mm}
GANs are powerful generative models that have shown state of the art performance in several tasks related to image and text generation. 
In this work, I use a variant of cGAN models to augment training data for low resource sentiment tasks.
Specifically, given Doc2Vec representation of sentences on datasets with a few training samples, I train a cGAN model using tricks such as pre-training and noise injection.
Followed by this, I generate fake feature vectors from the generator along with the associated sentiment class labels.
Empirically, I observe that augmenting classification with a model trained on the fake data provides gains in the low resource sentiment analysis tasks.
Further analysis shows that as the number of real samples for cGAN training increase, a baseline classifier trained on real data yields better accuracy in classifying the fake data. 
However, this does not translate to an equivalent increase in performance on real test instances when a classifier is trained on the fake data.  
Furthermore, I observe that fake data does not cover the entire region of feature space as occupied by real data using the t-SNE analysis.

In the future, I aim to address the observations made in the analysis of fake data distribution.
I aim to consider other techniques to obtain cGAN models that does not suffer from selective data generation limited to a smaller region of the feature space (as compared to the space spanned by real data).
Other GAN models that directly generate sentences are another attractive avenue for such a data augmentation.
The results of such as experiment would be more interpretable as one can directly observe the generated sentences. 
Finally, other variants of GAN model can be applied to the problem of sentiment analysis as well its extensions for this kind of data augmentation.

\begin{spacing}{0.95}
\bibliographystyle{IEEEbib}
\bibliography{strings,refs}
\end{spacing}

\end{document}